\documentclass[11pt]{article}
\usepackage{fontawesome5} 
\usepackage[final]{acl}
\usepackage{times}
\usepackage{latexsym}
\usepackage[T1]{fontenc}
\usepackage[utf8]{inputenc}
\usepackage{microtype}
\usepackage{inconsolata}
\usepackage{graphicx}
\usepackage{amsmath, amssymb}
\usepackage{multirow}  
\usepackage{graphicx}  
\usepackage{booktabs}  
\usepackage{makecell}
\usepackage[svgnames]{xcolor}
\newtheorem{theorem}{Theorem}

\newtheorem{proof}{Proof}[section]
\usepackage[table]{xcolor}

\title{ProFit: Leveraging High-Value Signals in SFT via Probability-Guided Token Selection}

\author{
Tao Liu$^{1,*}$ \quad
Taiqiang Wu$^{2,*,\dagger}$ \quad
Runming Yang$^{2}$ \quad
Shaoning Sun$^{1}$ \quad
Junjie Wang$^{1,\ddagger}$ \quad
Yujiu Yang$^{1,\ddagger}$
\\
$^{1}$Tsinghua University \quad
$^{2}$The University of Hong Kong
\\
\small{
$^{*}$Equal contribution \quad
$^{\dagger}$Project Leader \quad
$^{\ddagger}$Corresponding authors
}
\\
\tt\small{\href{https://github.com/Utaotao/ProFit}{\faGithub\ \texttt{https://github.com/Utaotao/ProFit}}}
}

\begin{document}
\maketitle
\begin{abstract}
Supervised fine-tuning (SFT) is a fundamental post-training strategy to align Large Language Models (LLMs) with human intent.
However, traditional SFT often ignores the \textit{one-to-many} nature of language by forcing alignment with a single reference answer, leading to the model overfitting to non-core expressions.
Although our empirical analysis suggests that introducing multiple reference answers can mitigate this issue, the prohibitive data and computational costs necessitate a strategic shift: prioritizing the mitigation of single-reference overfitting over the costly pursuit of answer diversity.
To achieve this, we reveal the intrinsic connection between token probability and semantic importance: high-probability tokens carry the core logical framework, while low-probability tokens are mostly replaceable expressions.
Based on this insight, we propose \textbf{ProFit}, which selectively masks low-probability tokens to prevent surface-level overfitting.
Extensive experiments confirm that ProFit consistently outperforms traditional SFT baselines on general reasoning and mathematical benchmarks.
\end{abstract}

\section{Introduction}

Large Language Models (LLMs) have demonstrated remarkable general capabilities~\citep{jaech2024openai,guo2025deepseek,yang2025qwen3}.
To adapt them to specific downstream tasks, Supervised Fine-Tuning (SFT) has become the prevailing paradigm~\citep{chung2024scaling}.
Traditional SFT is based on an autoregressive objective, forcing the model to strictly align with a single reference answer at the token level.
However, this rigid objective neglects the \textit{one-to-many} nature of language~\citep{li2016diversity,yang2025training}, where diverse expressions can convey the same intent.
Therefore, this strategy of forcibly fitting a single reference is often suboptimal and can easily lead to the model simply memorizing specific samples.~\citep{gudibande2023false,chu2025sft,wang2025think,wang2026perm}.

\begin{figure}[t]
\centering
\includegraphics[width=\linewidth]{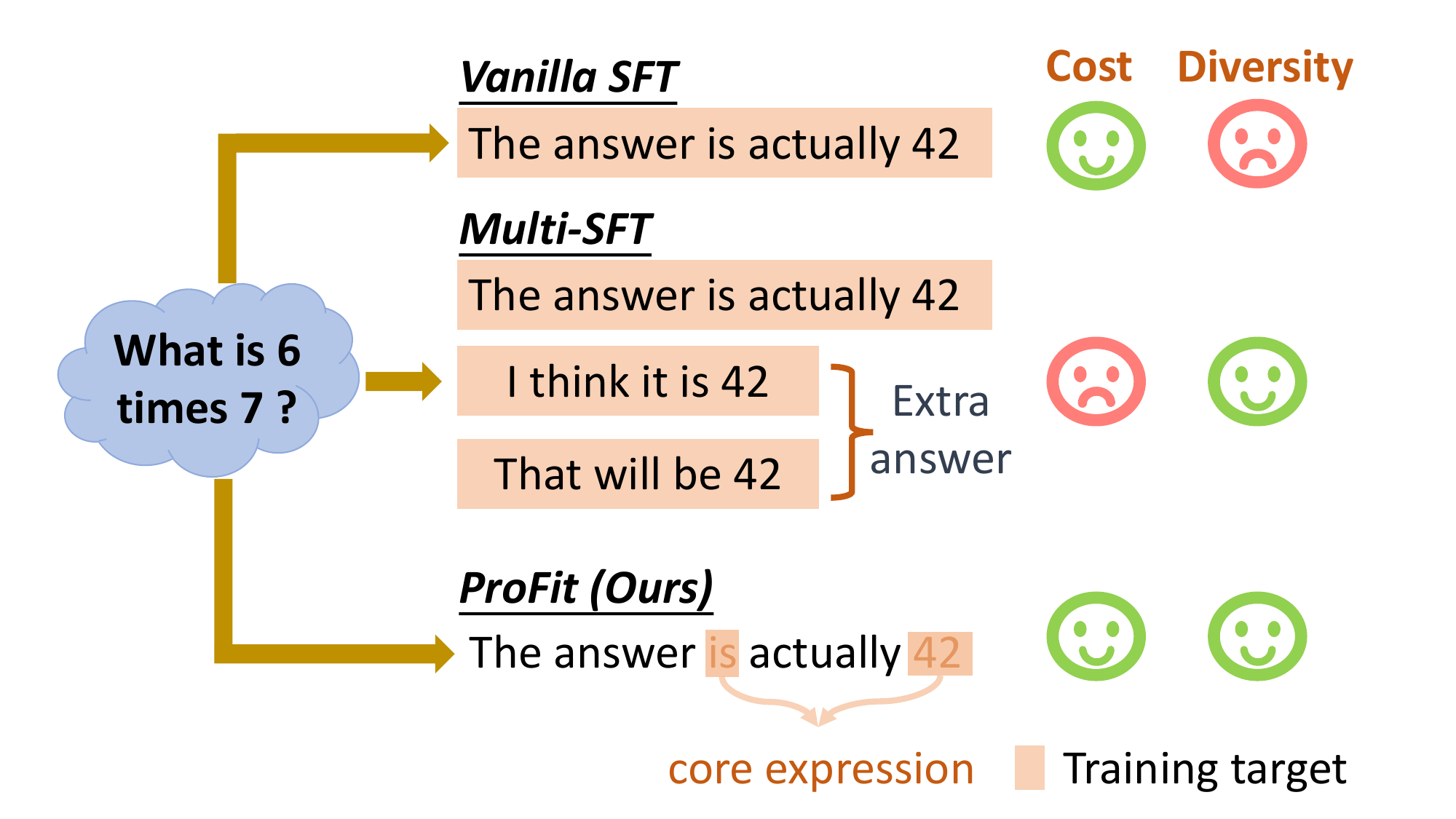}
\caption{\textbf{Breaking the trade-off between training cost and semantic diversity.} 
While Multi-reference SFT offers semantic richness at prohibitive data and computational costs, standard SFT is efficient but semantically limited.
ProFit achieves the best of both: by focusing supervision on high-value tokens, it captures core semantic integrity without sacrificing the efficiency of single-reference training.}
\label{fig:intro}
\end{figure}

While introducing multiple reference answers can alleviate this problem~\citep{yuan2023scalingrelationshiplearningmathematical,li2024preserving,shi2025reinforcement}, it faces the dual challenges of expensive data construction and difficulties in training convergence (please refer to Section~\ref{sec:Motivation} for details).
To solve this problem while maintaining the existing low-cost \textit{single instruction-single response} data configuration, we propose a more efficient strategy (as illustrated in Figure~\ref{fig:intro}): \textbf{instead of striving for comprehensive coverage with multiple answers, it's better to avoid overfitting to a single answer.}
To achieve this, we need a mechanism to filter out truly high-value training signals from a single reference answer.

To accurately identify these high-value training signals, we conducted a semantic analysis.
Specifically, we comparatively analyze multiple answers to a given question, aiming to identify those tokens defined as decisive for the answer's correctness, with their importance judged by Gemini-3-Pro.
Fortunately, we found that the predicted probability of the token can serve as an efficient and accurate proxy metric~\citep{kadavath2022language,huang2025low,bentegeac2025token}.
Further hypothesis testing confirmed this significant pattern: high-probability tokens tend to carry core reasoning logic or key semantics, while low-probability tokens correspond more to non-core expressions.

Inspired by this, we propose the \textbf{ProFit} method, which leverages the online probabilities predicted by the model currently being trained as the core clue to locate high-value signals.
Specifically, \textbf{ProFit} employs a strategic masking mechanism: it selectively retains and trains high-probability tokens that carry crucial semantic information, while masking low-probability, non-essential tokens~\citep{lin2024rho,ruan2025enhancing}.
We further provided theoretical derivations demonstrating that the gradients of low-probability tokens can overshadow the optimization direction of crucial tokens.

We conducted extensive evaluations on general reasoning (GPQA-Diamond~\cite{rein2024gpqa}), mathematics (MATH-500~\cite{lightman2023let}, AIME'24~\citep{aime2024dataset}, GSM8K~\cite{cobbe2021training}), and instruction following (IFEval~\cite{zhou2023instruction}).
The results consistently show that \textbf{ProFit} outperforms the traditional SFT baseline.
Notably, on the Qwen3 family, ProFit surpasses SFT by a significant margin of 3.0\% to 10.9\% in average accuracy, validating the effectiveness of our strategy.

Our contributions can be summarized as follows:
\begin{itemize}
    \item We identify a positive correlation between prediction probability and semantic importance, revealing that low-probability tokens typically represent non-essential expression.
    \item We propose \textbf{ProFit}, a probability-guided masking strategy. We theoretically prove that masking low-probability tokens prevents their large gradients from overshadowing key semantic signals.
    \item Extensive experiments on general reasoning and math benchmarks demonstrate that \textbf{ProFit} consistently outperforms standard SFT baselines.
\end{itemize}
\section{Related Work}

\subsection{Data-Efficient Instruction Tuning} 
Recent SFT adheres to the \textit{less is more} principle~\citep{zhou2023lima,zhang2025survey, li2025online,ji2026strideedstrategygroundedstepwisereasoning}, evolving from complexity-based selection~\citep{cao2023instruction, li2024superfiltering} to 2025's importance-aware metrics like MIWV~\citep{jiang2025importance} and ICL-based filtering~\citep{wang2025data, jiang2025importance}. 
However, these \textit{coarse-grained sample-level} methods treat pairs as atomic units, overlooking intra-sample low-information segments or stylistic noise~\citep{pang2025token, qin2025sstoken}, thus limiting the model's focus on dense logical signals.

\subsection{Token-Level Training Objectives} 
To address granularity limitations, research has pivoted to \textit{token-level} optimization~\citep{kong2025token}. 
While classical methods like Focal Loss~\citep{lin2017focal} and Unlikelihood Training~\citep{welleck2019neural} targeted hard tokens, recent LLM approaches like Rho-1~\citep{lin2024rho} and TIS-DPO~\citep{liu2024tis} rely on costly external reference models. 
More efficient intrinsic methods like DFT~\citep{wu2025generalization} employ probability-driven \textit{soft reweighting}, and CFT~\citep{ruan2025enhancing} further validates the need for supervising critical regions. 
Recently, \citet{li2025beyond} theoretically validated via a model-capability continuum that suppressing low-probability tokens benefits strong-prior domains like math, supporting our masking strategy.
While sharing the probability-driven motivation with DFT, \textbf{ProFit} diverges by adopting a strict \textit{hard masking} strategy to efficiently filter out non-core expressions in a single step.
\section{ Preliminaries and Motivation }

\subsection{Preliminaries}
\label{sec:Preliminaries}
\paragraph{Supervised Fine-Tuning.}
Given a dataset $\mathcal{D}$ containing pairs of inputs $x$ and reference responses $y^* = (y^*_1, \dots, y^*_T)$, SFT optimizes the policy $\pi_\theta$ by minimizing the negative log-likelihood:
\begin{equation}
    \mathcal{L}_{\mathrm{SFT}}(\theta) = \mathbb{E}_{(x,y^*)\sim \mathcal{D}} \left[ -\sum_{t=1}^{|y^*|} \log \pi_\theta(y^*_t \mid x, y^*_{<t}) \right].
    \label{SFT-loss}
\end{equation}
Let $z_t$ be the logits at step $t$. The probability is $p_{t} = \text{softmax}(z_t)$. The gradient of the per-token loss $\ell_t = -\log p_{t,y^*_t}$ with respect to the logits $z_t$ is:
\begin{equation}
    \frac{\partial \ell_t}{\partial z_{t,v}} = p_{t,v} - \mathbb{I}[v = y^*_t].
    \label{vacob gradient}
\end{equation}
Equation ~\ref{vacob gradient} drives $p_{t, y^*_t} \to 1$ while suppressing alternatives ($p_{t, v} \to 0$). 
This mechanism indiscriminately suppresses all non-reference tokens ($v \neq y^*_t$), including valid paraphrases, thereby penalizing semantic flexibility and driving the model toward \textit{surface-form overfitting}.

\paragraph{Low-Rank Adaptation (LoRA).}
LoRA is a parameter-efficient fine-tuning technique based on the premise that weight updates $\Delta W$ possess a low intrinsic rank~\citep{hu2022lora,wu2024mixture}. For a frozen weight matrix $W_0 \in \mathbb{R}^{d \times k}$, LoRA approximates the update via a low-rank decomposition $BA$, where $B \in \mathbb{R}^{d \times r}$ and $A \in \mathbb{R}^{r \times k}$ with $r \ll \min(d, k)$. The forward pass is modified as:
\begin{equation}
    h = W_0 x + \frac{\alpha}{r}\Delta W x = W_0 x + \frac{\alpha}{r}B A x
    \label{LoRA}
\end{equation}
where $\alpha$ is a scaling hyperparameter for adaptation stability.

\subsection{Motivation}
\label{sec:Motivation}

\begin{figure}[t]
\centering
\includegraphics[width=\linewidth]{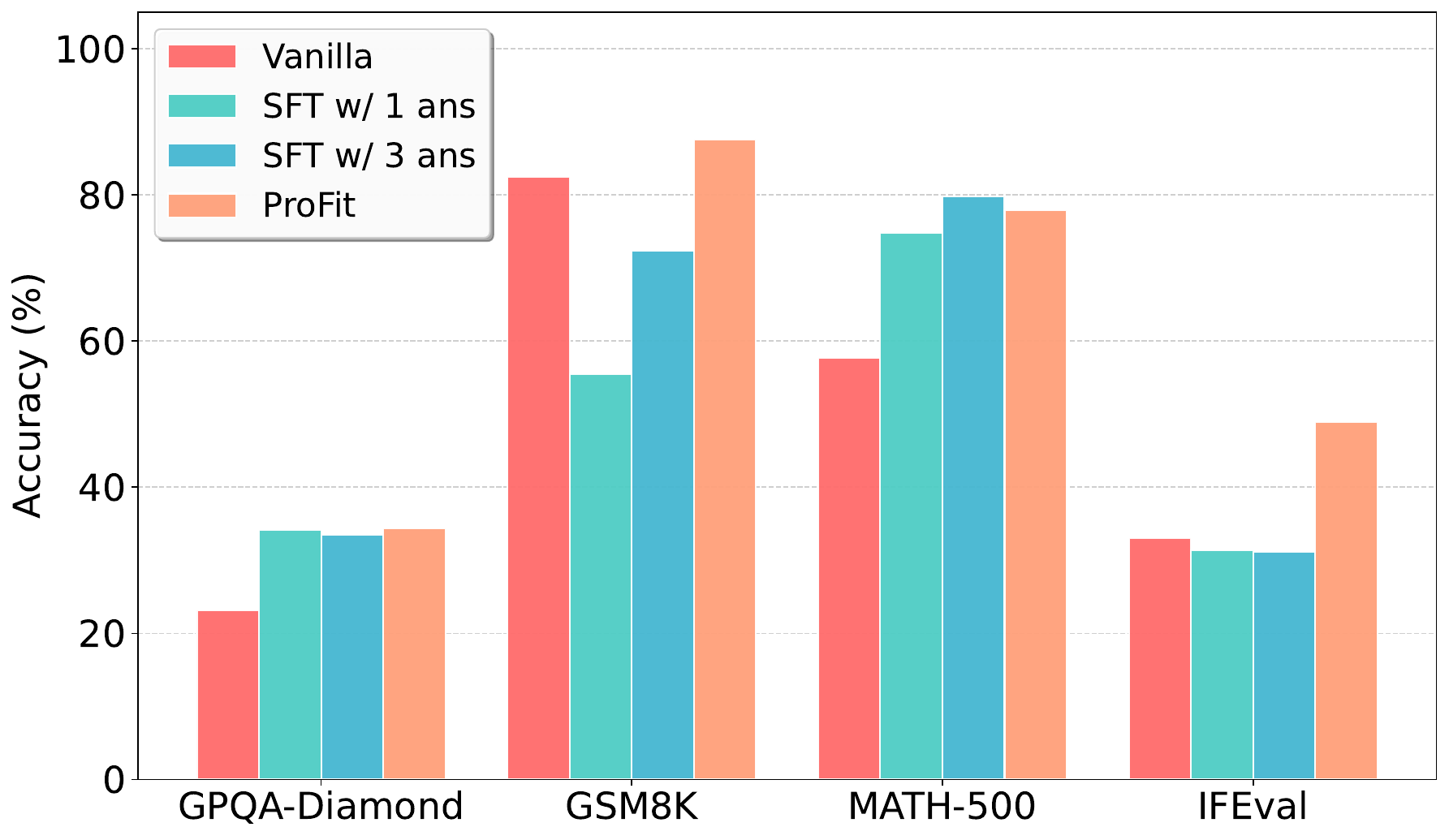}
\caption{\textbf{Performance comparison on diverse benchmarks.} 
While multi-reference training (SFT w/ 3 ans) offers sporadic gains, it suffers from optimization instability and stagnation on complex tasks. 
In contrast, \textbf{ProFit} achieves superior and robust performance across all metrics by selectively extracting high-value signals from a single reference.}
\label{fig:motivation}
\end{figure}

Traditional SFT relies on strict single-reference token-level alignment, which inherently over-penalizes paraphrastic variants by treating valid semantic equivalents as incorrect predictions. 
Intuitively, introducing multiple reference answers could theoretically bridge this gap. 
However, our pilot investigation reveals that this approach faces substantial practical barriers rather than offering a straightforward solution.

First, constructing a diverse, high-quality response set entails prohibitive costs: collecting $K$ distinct references per instruction scales the annotation burden linearly, and ensuring high quality often necessitates expert annotators, particularly for complex reasoning or mathematical tasks.
Second, and more critically, \textbf{simply expanding the reference set often introduces distributional conflicts, leading to optimization instability. }
As illustrated in Figure~\ref{fig:motivation}, we compared SFT trained with single versus multiple (3) reference answers. 
While multiple references yield marginal improvements on specific reasoning tasks like MATH-500, they fail to generalize consistently.
Surprisingly, on complex benchmarks like GPQA-Diamond, the performance stagnates or even slightly degrades (dropping from 34.1\% to 33.5\%) compared to the single-answer baseline. 
Furthermore, both standard and multi-answer SFT struggle to maintain base capabilities on instruction-following tasks like IFEval, where performance trails behind the Vanilla model.
This suggests that blindly fitting diverse distributions can confuse the model, causing it to struggle with convergence due to conflicting gradient directions.

To address this dilemma, we propose relaxing the strict alignment objectives. 
Instead of performing indiscriminate full-scale fitting on all tokens or relying on expensive multi-reference datasets, we implement a selective alignment strategy. 
\textbf{We need a mechanism that can accurately filter out the high-value training signals that truly carry the core reasoning logic from a single reference answer,} thereby achieving robust performance across diverse benchmarks.

\section{Methodology}
\label{}
\subsection{Semantic Analysis}
\label{Semantic Analysis}

To accurately extract high-quality training signals, we performed a joint analysis of semantic importance and prediction probability for the tokens in multiple reference answers.
Specifically, we utilize Gemini-3-pro~\citep{gemini3pro2025} as a semantic evaluator to annotate the tokens in the reference answers, classifying them into \textit{trivial tokens}, which represent interchangeable stylistic variations, and \textit{core tokens}, which encapsulate the essential reasoning logic.
Subsequently, we used Qwen3-4B-Base~\citep{yang2025qwen3} to perform forward propagation and calculate the predicted probability for each token. 
This choice provides a computation-efficient yet capable proxy for language modeling, striking a balance between estimation quality and inference cost.

\begin{figure}[t]
\centering
\includegraphics[width=\linewidth]{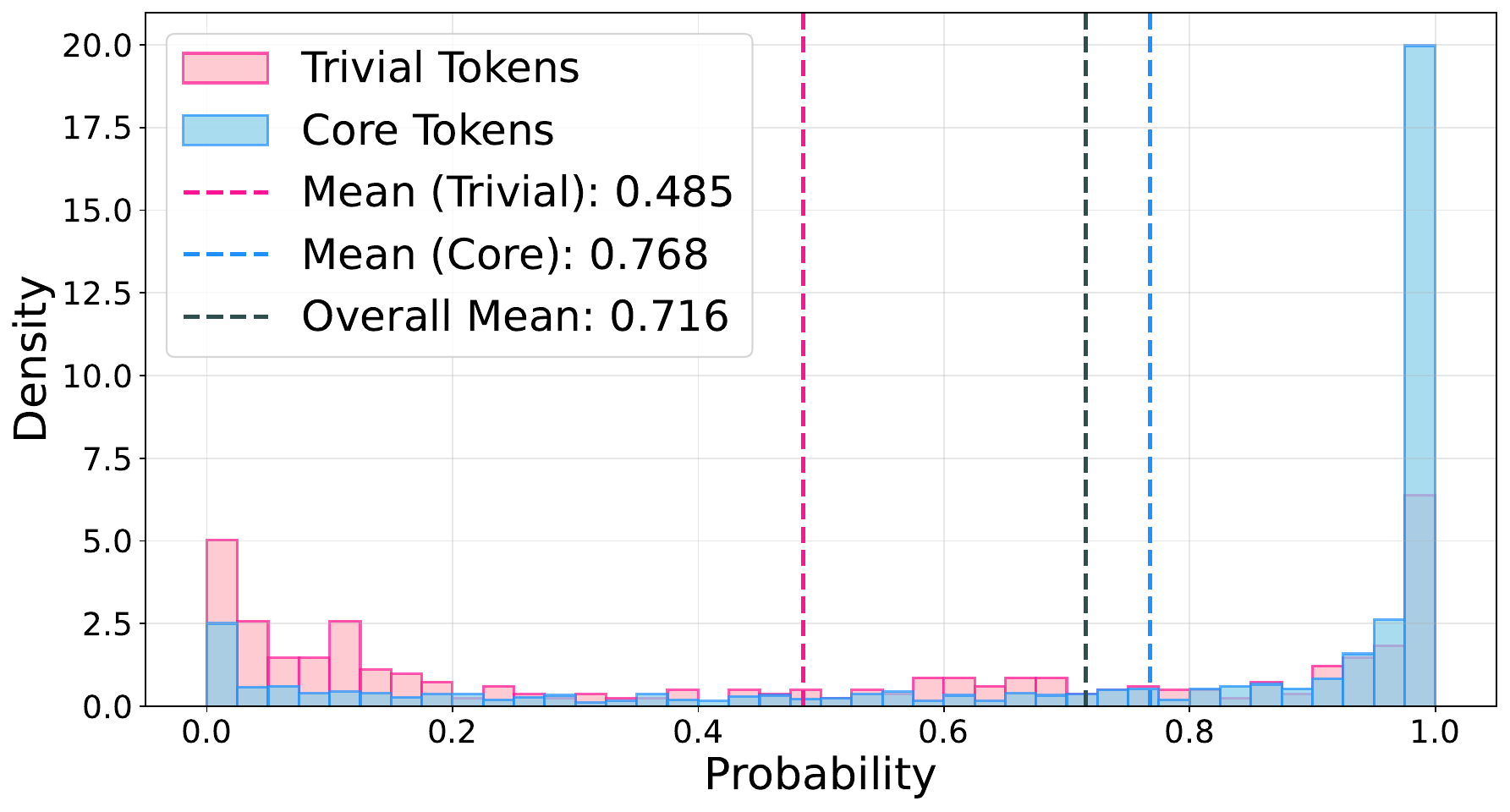}
\caption{Probability density estimation of semantic tokens. 
We categorize tokens into semantically \textit{\color{blue}{Core}} and \textit{\color{red!60}{Trivial}} groups. 
While \textit{core tokens} are heavily concentrated in high-confidence zones, \textit{trivial tokens} exhibit a significant \textbf{long-tail distribution}, disproportionately dominating the low-probability spectrum.
A hypothesis test confirms this significant distributional difference ($p = 1 \times 10^{-6}$).}
\label{fig:semantic_analysis}
\end{figure}

As shown in Figure~\ref{fig:semantic_analysis}, the two token types exhibit distinct probability distributions: core tokens are highly concentrated in the high-probability region, showing strong determinism, whereas trivial tokens display a long-tail distribution. 
Although some trivial tokens appear in the high-confidence interval, their density is much higher in the low-probability region than that of core tokens, making them the dominant component there.

To rigorously verify this observation, we conducted a statistical hypothesis test. We formally defined the null hypothesis as follows:
\begin{itemize}
    \item \textbf{Null Hypothesis ($H_0$):} The probability distributions of \textit{trivial tokens} and \textit{core tokens} are statistically identical (i.e., drawn from the same population).
    \item \textbf{Alternative Hypothesis ($H_1$):} The probability distributions of \textit{trivial tokens} and \textit{core tokens} are statistically distinct (i.e., drawn from different populations).
\end{itemize}
The test yielded a $p$-value of $1 \times 10^{-6}$, leading to a significant rejection of $H_0$.
This empirical result strongly supports our hypothesis: low prediction probability is a strong indicator of semantic non-essentiality, as low-probability regions are primarily dominated by \textit{trivial tokens}.

\subsection{ProFit}

Based on the motivation and semantic analysis, we propose \textbf{ProFit}.
The core intuition of this method is to utilize the model’s own prediction probabilities as a dynamic indicator to locate the core tokens.
During training, by implementing a threshold-based masking operator, ProFit selectively backpropagates gradients only from high-probability tokens, effectively isolating parameter updates from the interference of low-probability \textit{trivial tokens}.

To operationalize this strategy, we employ a \textbf{stop-gradient} mechanism to decouple the masking criterion from the gradient computation.
We define the binary validity mask $\mathcal{M}_t$ using the detached probability:
\begin{equation}
    \mathcal{M}_t = \mathbb{I}\left[ \text{sg}(\pi_{\theta}(y_{t}^* \mid x, y^*_{<t})) > \tau \right],
    \label{eq:mask}
\end{equation}
where $\text{sg}(\cdot)$ denotes the stop-gradient operator, $\tau \in [0, 1]$ is a static threshold, and $\mathbb{I}[\cdot]$ is the indicator function.
By strictly enforcing $p > \tau$, we ensure the optimization is driven solely by high-value semantic signals.
Crucially, the stop-gradient ensures that $\mathcal{M}_t$ acts as a fixed gate during backpropagation, avoiding the differentiability issues of the step function.

Formally, the optimization objective of ProFit is defined as:
\begin{equation}
    \mathcal{L}_{\mathrm{ProFit}}(\theta) = \mathbb{E}_{\mathcal{D}} \left[ - \frac{1}{T} \sum_{t=1}^{T} \mathcal{M}_t \log \pi_{\theta}(y_t^* \mid x, y^*_{<t}) \right], \label{eq:final_loss}
\end{equation}
where $T = |y^*|$ denotes the sequence length.

\subsection{Deeper Insights}
Equation~\ref{vacob gradient} shows that low-probability tokens induce significantly larger logit gradients. 
To quantify how this amplification propagates to the parameter space, we derive the following theorem, establishing a lower bound for the parameter gradient (proof in Appendix~\ref{sec:proof}):
\begin{theorem}
\label{theorem-grad}
\textbf{(Token-Wise Gradient Norm Lower Bound).} 
Consider the prediction of a single target token $y^*_t$ at step $t$, given the instruction $x$ and the preceding ground-truth tokens $y^*_{<t}$. 
Let $z \in \mathbb{R}^{|\mathcal{V}|}$ be the output logits and $\ell(\theta) = -\log \pi _\theta(y^*_t \mid x, y^*_{<t})$ be the loss for this step.
Let $J_\theta(z) = \nabla_\theta z \in \mathbb{R}^{|\mathcal{V}| \times \mid \theta \mid}$ denote the Jacobian of the logits with respect to parameters $\theta \in \mathbb{R}^{\mid \theta \mid}$.
Under the \textit{local non-degeneracy assumption} that the Jacobian is full row-rank (i.e., the model is locally surjective, satisfying $\sigma_{\min}(J_\theta(z)) \ge \gamma > 0$), the gradient norm satisfies:
\begin{equation}
    \|\nabla_\theta \ell\|_2 \ge \gamma \cdot (1 - \pi_\theta(y^*_t \mid x, y^*_{<t})).
\end{equation}
\end{theorem}
This lower bound theoretically guarantees that tokens with lower prediction probabilities inevitably induce larger parameter gradients.

\begin{table*}[!t]
\centering
\resizebox{\textwidth}{!}{
\begin{tabular}{ccccccc|cc} 
\toprule
\multicolumn{1}{c}{\textbf{Model}} & \textbf{Method} & \textbf{GPQA-Diamond} & \textbf{GSM8K} & \textbf{MATH-500} & \textbf{AIME'24} &  \textbf{IFEval} & \textbf{Avg.} & $\Delta$ \\
\midrule
\multirow{5}{*}{Qwen3-0.6B-Base } & Vanilla & 4.36 & 45.99 & 7.28 & 0.10 & 15.6 & 14.67 & -  \\
& SFT & 17.93 & 54.53 & 45.38 & 1.56 &23.15 & 28.51 & \textcolor{green!70!black}{+13.84} \\
& Entropy & 20.58 & 53.67 & 45.83 & 2.19 & 22.02 & 28.86 & \textcolor{green!70!black}{+14.19} \\
& DFT & 17.68 & 62.42 & 47.92 & 2.29 & 20.63 & 30.20 & \textcolor{green!70!black}{+15.53}\\
\rowcolor{blue!5} \cellcolor{white} & \textbf{ProFit} & 22.85 & 59.78 & 49.90 & 2.19 & 22.71 & \textbf{31.49} & \textcolor{green!70!black}{+16.82}\\
\midrule
\multirow{5}{*}{Qwen3-4B-Base} & Vanilla & 23.17 & 82.50 & 57.67 & 8.23 & 33.04 & 40.92 & - \\
 & SFT & 34.15 & 55.43 & 74.80 & 11.25 & 31.33 & 41.39 & \textcolor{green!70!black}{+0.47} \\
 & Entropy & 34.91 & 46.54 & 74.75 & 10.94 & 30.71 & 39.57 & \textcolor{red}{-1.35} \\
 & DFT & 31.69 & 87.83 & 77.50 & 10.83 & 43.67 & 50.30 & \textcolor{green!70!black}{+9.38} \\
 \rowcolor{ blue!5} \cellcolor{white} & \textbf{ProFit} & 34.34 & 87.55 & 77.85 & 13.02 & 48.87 & \textbf{52.33} & \textcolor{green!70!black}{+11.41} \\
 \midrule
 \multirow{5}{*}{Qwen3-14B-Base} & Vanilla & 37.69 & 83.44 & 78.32 & 13.33 & 52.61 & 53.08 & - \\
 & SFT & 46.02 & 78.00 & 79.22 & 16.04 & 36.69 & 51.20 & \textcolor{red}{-1.88} \\
 & Entropy & 47.85 & 78.23 & 80.12 & 14.79 & 38.01 & 51.80 & \textcolor{red}{-1.28} \\
 & DFT & 43.81 & 88.98 & 81.97 & 17.29 & 49.86 & 56.38 & \textcolor{green!70!black}{+3.30} \\
 \rowcolor{blue!5} \cellcolor{white} & \textbf{ProFit} & 46.53 & 89.62 & 82.85 & 16.56 & 58.02& \textbf{58.72} & \textcolor{green!70!black}{+5.64} \\
 \midrule
 \multirow{5}{*}{OLMo-2-7B} & Vanilla & 21.72 & 68.08 & 11.03 & 0.10 & 14.44 & 23.07 & - \\
 & SFT & 13.64 & 78.43 & 24.95 & 0.62 & 22.50 & 28.03 & \textcolor{green!70!black}{+4.96} \\
 & Entropy & 12.94 & 78.22 & 24.77 & 0.52 & 22.64 & 27.82 & \textcolor{green!70!black}{+4.75} \\
 & DFT & 12.31 & 76.09 & 23.57 & 0.31 & 23.24 & 27.10 & \textcolor{green!70!black}{+4.03} \\
 \rowcolor{blue!5} \cellcolor{white} & \textbf{ProFit} & 14.71 & 78.25 & 25.45 & 0.31 & 23.98 & \textbf{28.54} & \textcolor{green!70!black}{+5.47} \\
 \midrule
\multirow{5}{*}{Llama-3.1-8B} & Vanilla & 8.71 & 51.88 & 3.35 & 0.00 & 21.42 & 17.07 & - \\
 & SFT & 23.30 & 60.42 & 24.75 & 0.31 & 24.72 & 26.70 & \textcolor{green!70!black}{+9.63} \\
 & Entropy & 23.61 & 61.08 & 25.70 & 0.21 & 24.31 & 26.98 & \textcolor{green!70!black}{+9.91} \\
 & DFT & 8.40 & 60.07 & 17.65 & 0.62 & 25.14 & 22.38 & \textcolor{green!70!black}{+5.31} \\
 \rowcolor{blue!5} \cellcolor{white} & \textbf{ProFit} & 21.40 & 62.11 & 24.90 & 0.62 & 26.16 & \textbf{27.04} & \textcolor{green!70!black}{+9.97} \\

\bottomrule
\end{tabular}
}
\caption{Main results across five benchmarks. We report the accuracy of the Vanilla baseline, standard SFT, and varying strategies (Entropy, DFT, ProFit) on multiple model families. Regarding the evaluation settings, results on \textbf{AIME'24} are averaged over 32 samples, while results on the other datasets are averaged over 8 samples. The values in parentheses indicate the performance difference relative to the Vanilla baseline.}
\label{table:main_res}
\end{table*}
\section{Experiments}
\subsection{Experimental Setup}
\paragraph{Training.}
For the training data, we curated a subset of 2,000 samples from the BAAI-InfinityInstruct Dataset~\citep{zhou2023lima,li2025infinity,muennighoff2025s1}, prioritizing high reward scores as done in Shadow-FT~\citep{wu2025shadowfttuninginstructmodel}.
To comprehensively verify the effectiveness and generalization of our method, we conducted evaluations across a diverse set of LLMs, including the Qwen3 series~\citep{yang2025qwen3}, Llama 3 series~\citep{llama3modelcard}, and OLMo 2 series~\citep{olmo20242olmo2furious}. 
We employed LLaMA-Factory~\citep{zheng2024llamafactory} as our primary training framework. 
All experiments are conducted on 8 H20 GPUs.
For detailed hyperparameter settings, please refer to the Appendix~\ref{app:hyper-parameters}.
\paragraph{Baselines.}
To evaluate the effectiveness of ProFit, we compare it against several representative fine-tuning paradigms:
\begin{itemize}
    \item \textbf{Supervised Fine-tuning:} The vanilla baseline that minimizes the cross-entropy loss across all tokens indiscriminately.
    \item \textbf{Dynamic Fine-tuning}~\citep{wu2025generalization}\textbf{:} A probability-aware method that assigns a dynamic scale to the loss of each token based on its confidence.
    \item \textbf{Entropy-based tuning}~\citep{wang2025beyond}\textbf{:} A selective strategy that updates parameters only on tokens with high entropy.
\end{itemize}

\begin{figure*}[!t]
\centering
\includegraphics[width=\linewidth]{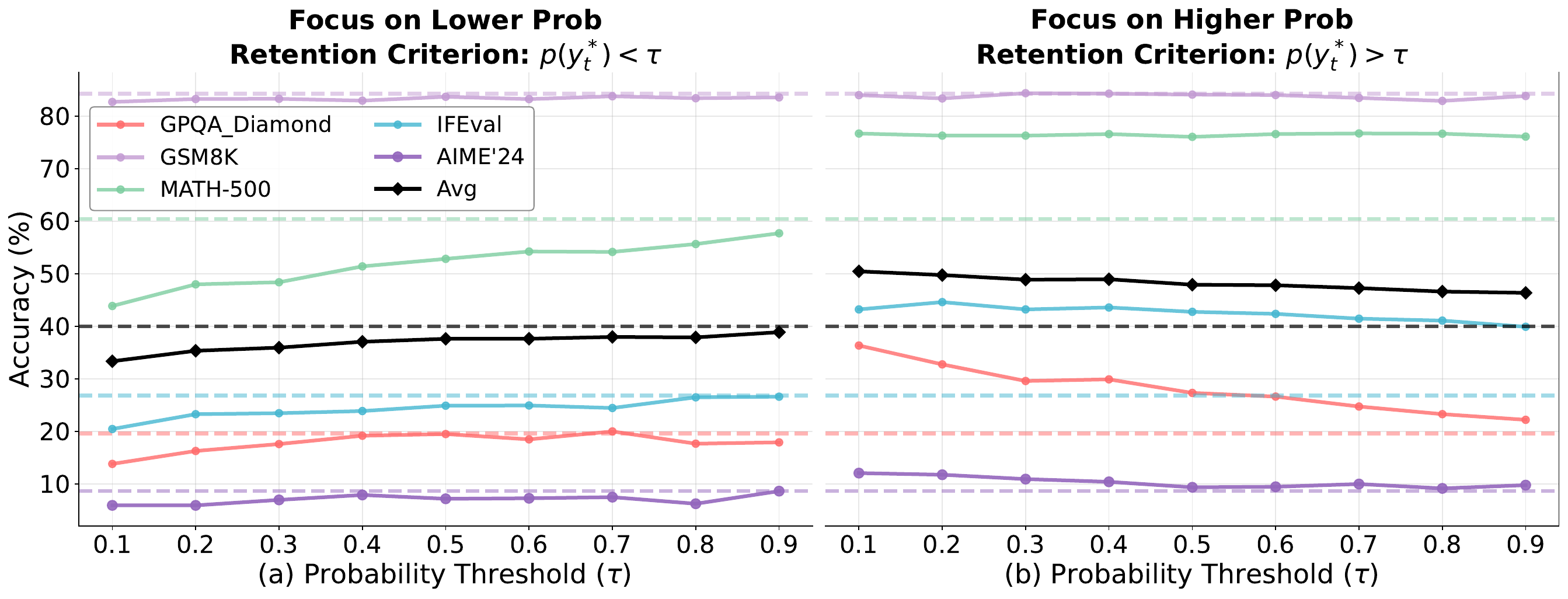}
\caption{Ablation study on the probability threshold $\tau$.
The \textbf{dashed line} represents the performance of the standard SFT baseline.
(a) Training exclusively on low-probability tokens ($p(y_t^*) < \tau$) results in performance consistently below the baseline, indicating that non-core expressions are insufficient for constructing effective reasoning chains.
(b) Conversely, the proposed strategy ($p(y_t^*) > \tau$), which masks low-probability noise, consistently outperforms the baseline across all tasks, validating the effectiveness of focusing on core logic.}
\label{fig:threshold}
\end{figure*}

\paragraph{Evaluation.}
To comprehensively evaluate the downstream performance of our fine-tuned models, we conducted extensive assessments across a diverse set of benchmarks, including GPQA-Diamond~\citep{rein2024gpqa}, MATH-500~\citep{lightman2023let}, GSM8K~\citep{cobbe2021training}, AIME'24~\citep{aime2024dataset}, and IFEval~\citep{zhou2023instruction}. 
For the evaluation pipeline, we utilized OpenCompass~\citep{2023opencompass} as the primary inference framework, integrated with lmdeploy~\citep{2023lmdeploy} and vllm~\citep{kwon2023efficient} as the acceleration backend. 
For the decoding strategy, we performed sampling and reported the average accuracy across 32 generations for AIME'24 and 8 for other benchmarks.
Detailed inference hyperparameters and configuration settings are provided in the Appendix~\ref{app:hyper-parameters}.

\subsection{Main Results}
Table~\ref{table:main_res} presents the comparative results of our proposed method, ProFit, against the Vanilla baseline, standard SFT, and other strategies (Entropy and DFT) across five diverse benchmarks. 
The experimental results demonstrate that ProFit consistently achieves superior performance across all evaluated model families and scales.

\textbf{Compared to standard SFT, ProFit delivers substantial improvements in average accuracy. }
For instance, on the Qwen3-4B-Base model, ProFit achieves an average accuracy of 52.33\%, surpassing standard SFT (41.39\%) by a significant margin of 10.94\%. 
Similarly, on Llama-3.1-8B, ProFit improves the average score to 27.04\%, outperforming SFT (26.70\%) and showing a +9.97\% gain over the Vanilla baseline. 
Notably, on Qwen3-14B-Base, standard SFT experiences a performance drop of $1.88\%$ relative to Vanilla, likely stemming from the interference of non-core expressions or superficial stylistic patterns. 
ProFit successfully mitigates this issue, reversing the decline to achieve a distinct gain of $+5.64\%$.
This underscores the stability of our probability-guided filtering mechanism in mitigating negative transfer while enhancing downstream performance.

\textbf{ProFit also consistently outperforms other baselines.}
As shown in Table~\ref{table:main_res}, while DFT and Entropy strategies generally offer improvements over the Vanilla baseline, they often fall short of the gains achieved by ProFit. 
For example, on Qwen3-0.6B, ProFit achieves the highest average accuracy of 31.49\%, exceeding DFT (30.20\%) and Entropy (28.86\%). 
This trend holds across architectures like OLMo-2 and Llama-3.1, showing our probability-based identification of non-core expressions outperforms entropy-based filtering and dynamic reweighting methods.

In summary, ProFit shows strong scalability and universality, consistently boosting performance across model sizes and architectures without the high data costs of multi-reference fine-tuning.

\section{Extensive Analysis}
\begin{figure}[!t]
\centering
\includegraphics[width=\linewidth]{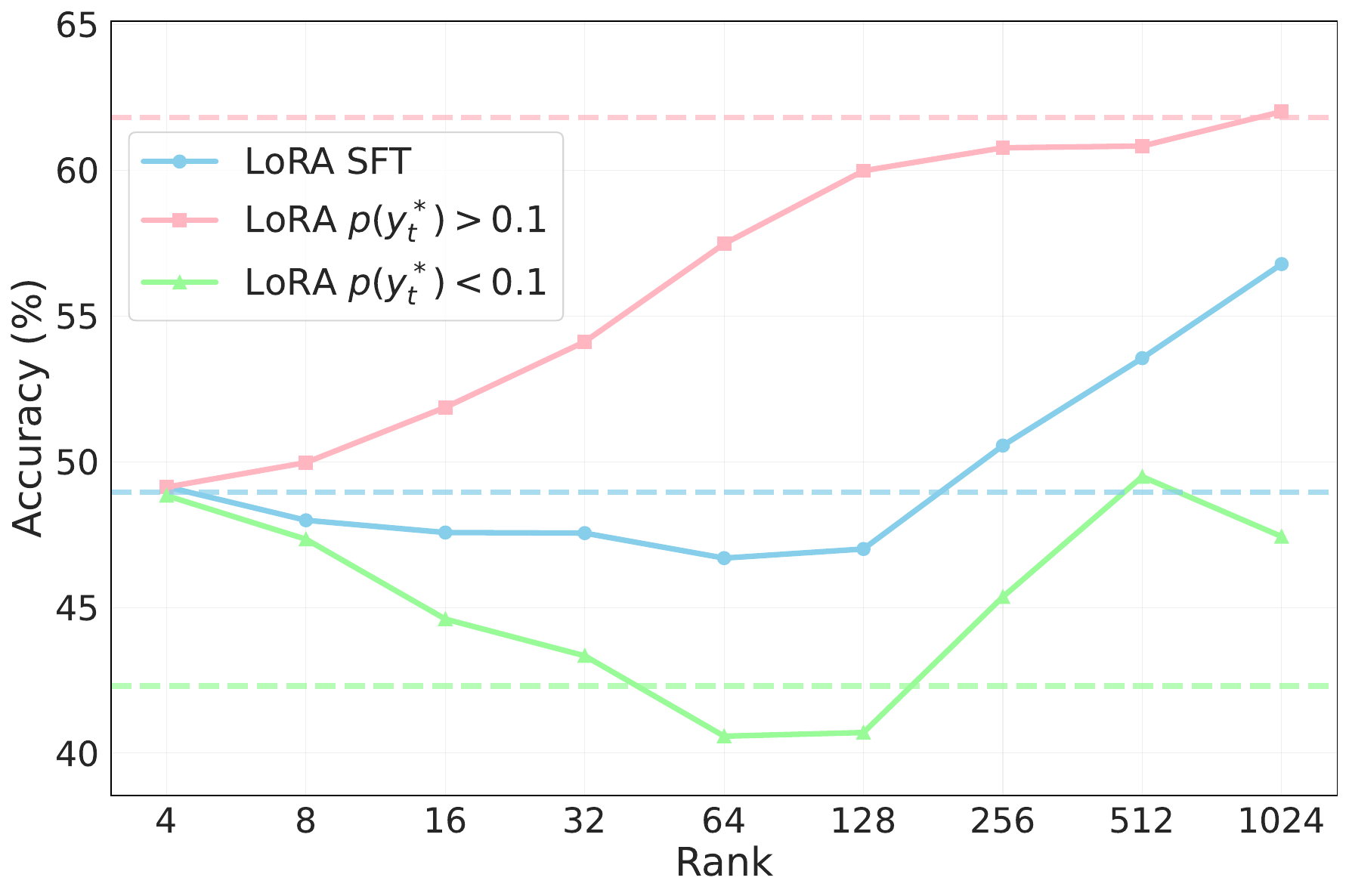}
\caption{Average performance variation across different LoRA ranks ($r \in \{4, \dots, 1024\}$). The dashed lines represent the baseline performance of full-parameter fine-tuning for each corresponding setting. While core tokens ($p(y_t^*) > 0.1$) exhibit monotonic improvement driven by capacity, non-core tokens ($p(y_t^*) < 0.1$) and standard SFT show a U-shaped trend, revealing optimization interference at medium ranks.}
\label{fig:LoRA_rank}
\end{figure}

\subsection{Analysis of \texorpdfstring{$\tau$}{tau} parameter}
\label{sec:ablation_threshold}
To investigate the impact of the probability threshold $\tau$, we compared the experimental results of retaining low-probability tokens ($p(y_t^*) < \tau$) versus retaining high-probability tokens ($p(y_t^*) > \tau$), as illustrated in Figure \ref{fig:threshold}.
 
The Figure \ref{fig:threshold}(b) reveals that when masking low-probability tokens—which typically represent semantic diversity—the model's performance consistently exceeds the full-token fine-tuning baseline (dashed line) across all threshold settings. 
This suggests that in traditional SFT, forcing the model to fit these surface-level stylistic variations distracts it from learning the underlying reasoning patterns.
By alleviating this unnecessary learning burden, the model can focus more effectively on the invariant logical core. 
Although knowledge-intensive tasks like GPQA-Diamond show a slight performance decline as the threshold increases, likely due to specific long-tail entities falling into the low-probability range, their absolute performance remains significantly above the baseline.

Conversely, the Figure \ref{fig:threshold}(a) highlights the irreplaceable skeletal role of high-probability tokens. 
When the model is restricted to learning only low-probability \textit{diverse expressions} without the support of high-probability structural tokens, performance suffers a catastrophic decline. 
Crucially, even as the threshold $\tau$ increases (introducing more high-frequency tokens), while performance recovers slightly, it never reaches the level of the full-token fine-tuning baseline and remains far inferior to the strategy shown in the \ref{fig:threshold}(b).
This persistent performance gap indicates that low-probability tokens are essentially auxiliary components contingent upon the logical skeleton. 
Without the support of high-confidence core logic, merely increasing the fitting of these non-core expressions fails to establish effective reasoning links and ultimately limits the model's generalization potential.

\subsection{Impact of LoRA Rank}

We explore the influence of trainable parameter volume by scaling the LoRA rank from 4 to 1024, as shown in Figure~\ref{fig:LoRA_rank} (refer to Appendix~\ref{app:rank_details} for details).
The results reveal a distinct divergence: core tokens ($p > 0.1$) benefit monotonically from increased rank, indicating they strictly require model capacity.
In contrast, trivial tokens ($p < 0.1$) and standard SFT display a U-shaped trend, where medium ranks struggle with optimization interference.
Interestingly, for trivial tokens, LoRA (Rank 1024) outperforms full fine-tuning, demonstrating that low-rank constraints serve as effective regularization, preventing the model from overfitting to non-essential statements.
The fact that global SFT follows the trivial token trend underscores that non-core expressions act as the primary bottleneck in standard training.

\subsection{Performance Evolution across Epochs}

\begin{figure}[!t]
\centering
\includegraphics[width=\linewidth]{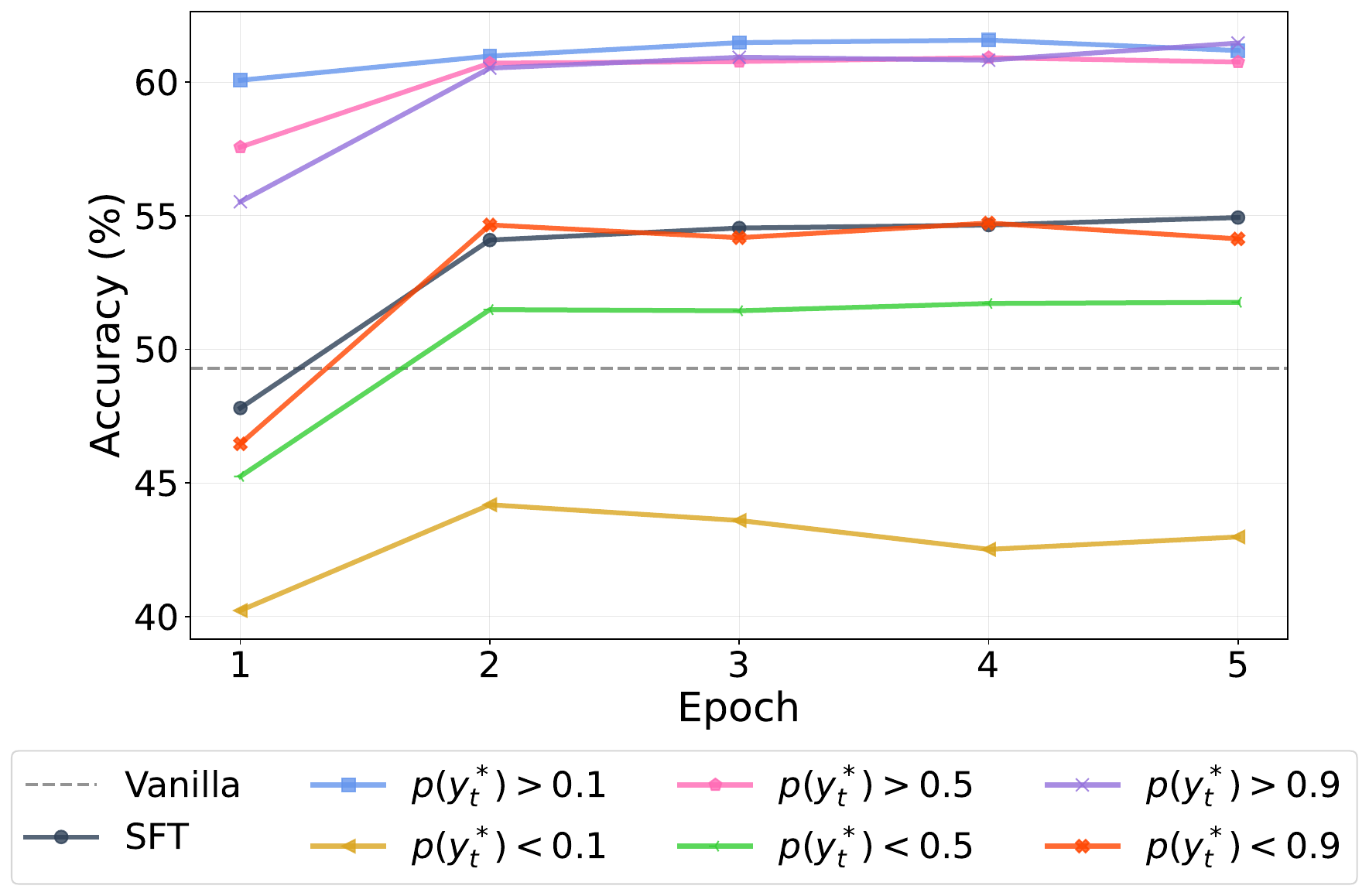}
\caption{Average performance trajectory across training epochs. ProFit ($p > \tau$) demonstrates rapid convergence and a superior performance ceiling, whereas focusing on low-probability tokens ($p < \tau$) results in training instability and limited capacity.}
\label{fig:epoch}
\end{figure}

Figure \ref{fig:epoch} illustrates the training trajectory over 5 epochs.
ProFit ($p > \tau$) demonstrates superior efficiency, converging immediately to 60.1\% accuracy in the first epoch—already surpassing the Baseline's peak performance of 54.9\%.
In stark contrast, training exclusively on low-probability tokens ($p < \tau$) leads to stagnation in a suboptimal range ($40\%-50\%$) and training instability.
These results confirm that high-probability tokens contain the essential gradient signals for alignment, whereas low-probability regions offer negligible or even detrimental supervision.

\begin{figure*}[!t]
\centering
\includegraphics[width=\linewidth]{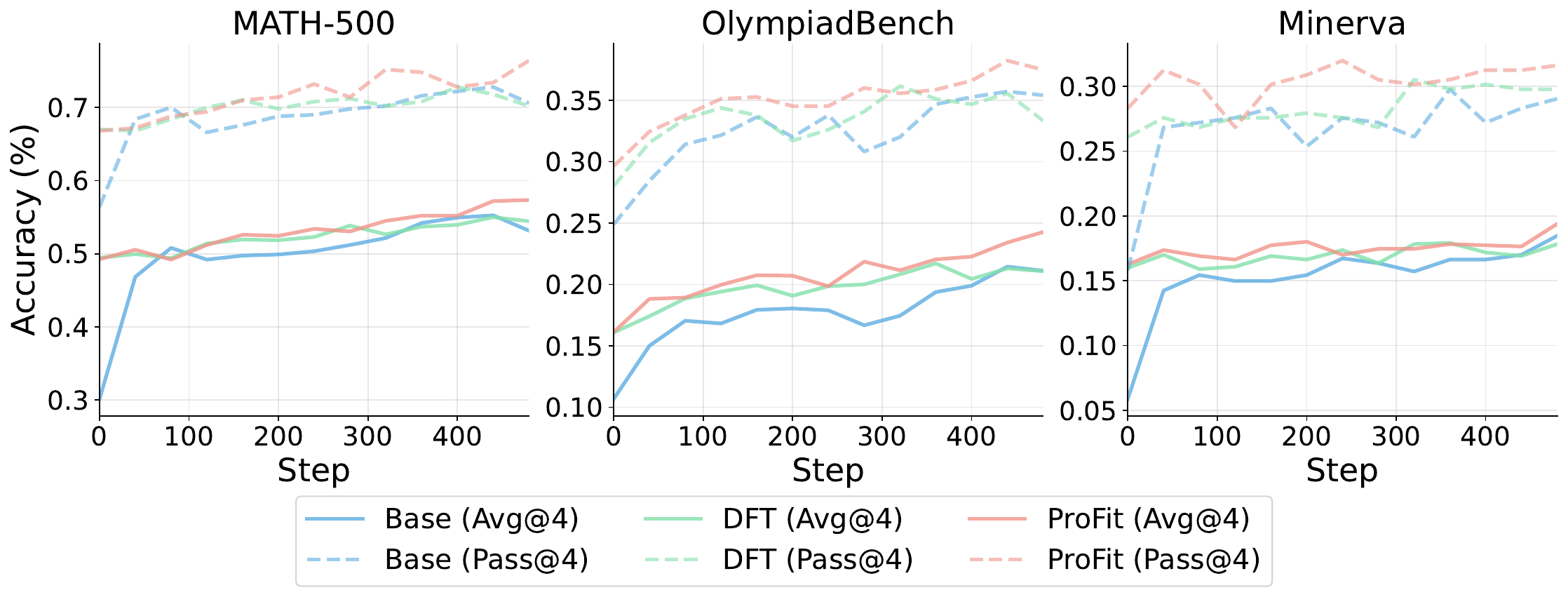}
\caption{Performance trajectories on MATH-500, OlympiadBench, and Minerva datasets. ProFit combines high initial performance with continuous learning capability, ultimately achieving the best Pass@4 and  Avg@4 scores across all benchmarks.}
\label{fig:rl_results}
\end{figure*}

\subsection{Superior RL Initialization}
We conduct experiments based on the Qwen3-0.6B-Base model. 
Specifically, we employ GRPO (Group Relative Policy Optimization~\citep{shao2024deepseekmath}) to optimize the models initialized with three different strategies: Base, DFT, and our proposed ProFit.
We employ the DeepScaleR dataset as training prompts following \citet{liu2025learnreasonefficientlyadaptive}.
All experiments are conducted on 32 H20 GPUs. 
For evaluation, in addition to the previously introduced MATH-500, we employ Minerva~\citep{hendrycks2021measuring} and OlympiadBench~\citep{he2024olympiadbench}, which serve as comprehensive benchmarks covering tasks from general mathematical reasoning to diverse competition-level challenges.

We evaluated the potential of ProFit as an initialization for subsequent reinforcement learning by analyzing its training dynamics across three mathematical benchmarks, as illustrated in Figure \ref{fig:rl_results}. 
The experimental results demonstrate that ProFit consistently outperforms comparison methods, offering not only a superior starting point but also more robust training stability for the RL stage.
Please refer to Appendix~\ref{app:appendix_rl_dynamics} for detailed training dynamics, such as KL divergence, entropy, and response length.

Specifically, on the MATH-500 dataset, ProFit achieves an Avg@4 of \textbf{57.3\%} and a Pass@4 of \textbf{76.4\%} at the final training stage, significantly surpassing the Baseline which reaches 53.1\% and 70.6\% respectively. 
Crucially, the analysis of training curves reveals distinct convergence behaviors: while the Baseline exhibits a pronounced \textit{cold start} phenomenon requiring more steps to learn effective patterns, and DFT shows earlier plateauing in Avg@4 under the same budget, ProFit maintains continuous performance growth throughout the training process. 

This advantage is particularly evident on the challenging OlympiadBench, where ProFit attains an Avg@4 of \textbf{24.3\%}, distinctively outperforming the competing methods which hover around the 21.1\% level. 
Similar consistent improvements are also observed on Minerva, further validating the generalization capability of our method.

\subsection{Analysis of Fine-Tuning Instruct Models}
Further fine-tuning instruction-aligned models typically incurs an alignment tax, degrading general capabilities~\cite{wu2025shadowfttuninginstructmodel,wu2025generalization,diao2026entropy}. We evaluate ProFit on the Qwen2.5-Instruct series to test its robustness against this issue. As shown in Table \ref{table:instruct_models_appendix}, while standard SFT and DFT consistently suffer performance drops (e.g., -1.83 average degradation for SFT on 7B), ProFit effectively mitigates catastrophic forgetting. It not only prevents typical declines but actually improves the average scores for the 0.5B (+1.24) and 7B (+0.17) models, with notable gains in reasoning tasks like GSM8K and AIME'24, proving its superiority in refining aligned models without sacrificing foundational instruction-following skills.

\begin{table*}[!t]
\centering
\resizebox{\textwidth}{!}{
\begin{tabular}{ccccccc|cc} 
\toprule
\multicolumn{1}{c}{\textbf{Model}} & \textbf{Method} & \textbf{GPQA-Diamond} & \textbf{GSM8K} & \textbf{MATH-500} & \textbf{IFEval} & \textbf{AIME'24} & \textbf{Avg.} & $\Delta$ \\
\midrule
\multirow{4}{*}{Qwen2.5-0.5B-Instruct} & Vanilla & 17.42 & 37.38 & 27.98 & \textbf{25.28} & 0.21 & 21.65 & -  \\
& SFT & \textbf{17.55} & 37.96 & 24.47 & 23.66 & 0.21 & 20.77 & \textcolor{red}{-0.88} \\
& DFT & 10.67 & 40.48 & 27.65 & 25.25 & 0.31 & 20.87 & \textcolor{red}{-0.78} \\
\rowcolor{blue!5} \cellcolor{white} & \textbf{ProFit} & 16.60 & \textbf{42.97} & \textbf{29.55} & 24.72 & \textbf{0.62} & \textbf{22.89} & \textcolor{green!70!black}{+1.24}\\
\midrule
\multirow{4}{*}{Qwen2.5-3B-Instruct} & Vanilla & \textbf{32.01} & 74.73 & 65.78 & \textbf{58.99} & 4.38 & \textbf{47.18} & - \\
 & SFT & 30.30 & 75.29 & 63.52 & 57.44 & 5.31 & 46.37 & \textcolor{red}{-0.81} \\
 & DFT & 26.58 & 77.27 & 65.77 & 50.72 & \textbf{7.60} & 45.59 & \textcolor{red}{-1.59} \\
 \rowcolor{ blue!5} \cellcolor{white} & \textbf{ProFit} & 29.10 & \textbf{77.90} & \textbf{65.83} & 52.22 & 7.19 & 46.45 & \textcolor{red}{-0.73} \\
 \midrule
 \multirow{4}{*}{Qwen2.5-7B-Instruct} & Vanilla & \textbf{35.48} & 81.00 & \textbf{75.28} & \textbf{69.29} & 12.50 & 54.71 & - \\
 & SFT & 33.27 & 83.82 & 74.65 & 61.95 & 10.73 & 52.88 & \textcolor{red}{-1.83} \\
 & DFT & 32.77 & 85.60 & 74.65 & 60.28 & 13.02 & 53.26 & \textcolor{red}{-1.45} \\
 \rowcolor{blue!5} \cellcolor{white} & \textbf{ProFit} & 35.35 & \textbf{85.61} & 75.15 & 63.79 & \textbf{14.48} & \textbf{54.88} & \textcolor{green!70!black}{+0.17} \\
\bottomrule
\end{tabular}
}
\caption{Performance evaluation on previously instruction-tuned models (Qwen2.5-Instruct series). We report the accuracy of the original Instruct model (Vanilla), standard SFT, DFT, and ProFit. }
\label{table:instruct_models_appendix}
\end{table*}
\begin{table*}[!t]
\centering
\resizebox{\textwidth}{!}{
\begin{tabular}{ccccccc|cc} 
\toprule
\multicolumn{1}{c}{\textbf{Model}} & \textbf{Method} & \textbf{HellaSwag} & \textbf{ARC-c} & \textbf{HumanEval} & \textbf{MMLU-Pro History} &  \textbf{MMLU-Pro Chemistry} & \textbf{Avg.} & $\Delta$ \\
\midrule
\multirow{5}{*}{Qwen3-0.6B-Base} & Vanilla & 24.54 & 1.69 & 1.45 & 1.31 & 3.18 & 6.43 & -  \\
& SFT & 33.37 & 45.76 & 32.09 & 15.22 & 15.72 & 28.43 & \textcolor{green!70!black}{+22.00} \\
& Entropy & 33.16 & 50.51 & 31.78 & 11.02 & 15.90 & 28.47 & \textcolor{green!70!black}{+22.04} \\
& DFT & 32.72 & 53.22 & 37.80 & 10.76 & 14.93 & 29.89 & \textcolor{green!70!black}{+23.46}\\
\rowcolor{blue!5} \cellcolor{white} & \textbf{ProFit} & \textbf{33.80} & \textbf{55.59} & \textbf{38.64} & \textbf{17.32} & \textbf{19.61} & \textbf{32.99} & \textcolor{green!70!black}{+26.56}\\
\midrule
\multirow{5}{*}{Qwen3-4B-Base} & Vanilla & 65.55 & 34.92 & 75.38 & 27.03 & 16.96 & 43.97 & - \\
 & SFT & \textbf{74.83} & \textbf{92.20} & \textbf{82.85} & 41.99 & \textbf{62.46} & \textbf{70.87} & \textcolor{green!70!black}{+26.90} \\
 & Entropy & 74.77 & 90.85 & 81.78 & 41.73 & 61.31 & 70.09 & \textcolor{green!70!black}{+26.12} \\
 & DFT & 62.12 & 90.85 & 81.40 & 42.26 & 59.28 & 67.18 & \textcolor{green!70!black}{+23.21} \\
 \rowcolor{ blue!5} \cellcolor{white} & \textbf{ProFit} & 72.21 & 91.19 & 80.79 & \textbf{44.62} & 61.04 & 69.97 & \textcolor{green!70!black}{+26.00} \\
 \midrule
 \multirow{5}{*}{Qwen3-14B-Base} & Vanilla & 84.02 & \textbf{94.92} & 83.69 & 54.59 & 57.95 & 75.03 & - \\
 & SFT & 82.18 & 94.58 & 86.97 & 53.81 & 70.05 & 77.52 & \textcolor{green!70!black}{+2.49} \\
 & Entropy & 81.43 & 93.90 & 87.04 & 55.38 & \textbf{70.41} & 77.63 & \textcolor{green!70!black}{+2.60} \\
 & DFT & 86.96 & 94.58 & 87.65 & 56.43 & 69.70 & 79.06 & \textcolor{green!70!black}{+4.03} \\
 \rowcolor{blue!5} \cellcolor{white} & \textbf{ProFit} & \textbf{87.30} & \textbf{94.92} & \textbf{89.18} & \textbf{59.84} & 70.32 & \textbf{80.31} & \textcolor{green!70!black}{+5.28} \\
\bottomrule
\end{tabular}
}
\caption{Performance comparison on General Tasks across different Qwen3 model scales (0.6B, 4B, 14B). We report the accuracy of the Vanilla baseline, standard SFT, and varying fine-tuning strategies (Entropy, DFT, ProFit). }
\label{table:general_tasks_appendix}
\end{table*}

\subsection{Analysis of Performance on General Tasks}
To evaluate whether ProFit preserves comprehensive abilities while optimizing for specific domains, we tested it on general benchmarks across Qwen3 scales. As shown in Table \ref{table:general_tasks_appendix}, ProFit demonstrates strong generalization without severe catastrophic forgetting. It achieves the highest average improvements on the 0.6B (+26.56) and 14B scales over the Vanilla model. Even on the 4B scale, where standard SFT slightly edges out in the overall average, ProFit remains highly competitive and excels in specific reasoning-heavy subsets , proving it effectively balances broad knowledge retention with its primary fine-tuning objectives.

\section{Conclusion}

In this work, we propose ProFit, a novel method to fine-tune LLMs by leveraging token prediction probabilities as a proxy for semantic importance. 
Inspired by our hypothesis testing results revealing that high-probability tokens carry core semantics while low-probability ones correspond to non-core expressions, we propose ProFit to selectively mask the latter, aiming to alleviate the overfitting to surface-level phrasing and capture the underlying logic.
Extensive experiments across multiple LLM series, including Qwen, Llama, and OLMo2, demonstrate that ProFit consistently outperforms conventional full-parameter SFT and other data selection methods on diverse benchmarks covering general reasoning, mathematics, and instruction following. 
We further conduct comprehensive ablation studies on training epochs, probability thresholds, and LoRA ranks. 
ProFit also serves as a superior initialization for reinforcement learning, promoting stable convergence and deeper reasoning exploration.
These analyses confirm that low-probability tokens serve as a primary source of optimization interference, validating that our approach offers a robust and effective solution for improving model generalization.

\section*{Acknowledgements}
This work was partly supported by the National Natural Science Foundation of China (Grant No. 62576191), the Shenzhen Science and Technology Program (ZDCY20250901103533010) and Tsinghua SIGS KA Cooperation Fund.

\section*{Limitation}

Despite the promising results, this work has limitations.

First, our core assumption that low-probability tokens represent non-core expressions primarily holds for logic-intensive tasks (e.g., reasoning, mathematics). 
For creative generation tasks, such tokens may contribute to stylistic diversity, which requires further investigation. 

Second, ProFit currently employs a static probability threshold across all samples. 
While this design choice was made to ensure implementation simplicity and training stability and has yielded consistent empirical gains, we acknowledge that future iterations could further enhance performance by exploring adaptive mechanisms that dynamically adjust the threshold based on instance-specific difficulty.

\clearpage
\appendix
\section*{Appendix}
\section{Proof of Theorem~\ref{theorem-grad}}
\label{sec:proof}

\begin{proof}
Consider the loss $\ell(\theta)$ for the target token $y^*_t$ conditioned on the input $x$ and history $y^*_{<t}$. Let $z \in \mathbb{R}^{|\mathcal{V}|}$ be the logit vector and $p = \text{softmax}(z)$ be the probability distribution.
Recall from Eq.~\ref{vacob gradient} that the gradient of the loss with respect to the logits is $\nabla_z \ell = p - \mathbf{e}_{y^*_t}$, where $\mathbf{e}_{y^*_t}$ is the one-hot vector corresponding to the target.

First, we analyze the $\ell_2$-norm of the logit gradient $\nabla_z \ell$. Let $p_{y^*_t} = p_\theta(y^*_t \mid x, y^*_{<t})$ denote the predicted probability of the ground-truth token. We derive:
\begin{equation}
\begin{aligned}
    \|\nabla_z \ell\|_2 
    &= \sqrt{(p_{y^*_t} - 1)^2 + \sum_{v \neq y^*_t} p_v^2} \\
    &= \sqrt{(1 - p_{y^*_t})^2 + \sum_{v \neq y^*_t} p_v^2} \\
    &\ge 1 - p_{y^*_t}.
\end{aligned}
\label{eq:logit-grad-norm}
\end{equation}

Next, applying the chain rule, the gradient with respect to the model parameters $\theta$ is given by:
\begin{equation}
    \nabla_\theta \ell = \left(\frac{\partial z}{\partial \theta}\right)^\top \nabla_z \ell = J_\theta(z)^\top \nabla_z \ell,
\end{equation}
where $J_\theta(z) \in \mathbb{R}^{|\mathcal{V}| \times \mid \theta \mid}$ is the Jacobian matrix.
We utilize the spectral inequality for matrix-vector products: for any matrix $\mathbf{A}$ and vector $\mathbf{v}$, $\|\mathbf{A}\mathbf{v}\|_2 \ge \sigma_{\min}(\mathbf{A}) \|\mathbf{v}\|_2$. Applying this to our context:
\begin{equation}
\begin{aligned}
    \|\nabla_\theta \ell\|_2 
    &= \|J_\theta(z)^\top \nabla_z \ell\|_2 \\
    &\ge \sigma_{\min}(J_\theta(z)^\top) \cdot \|\nabla_z \ell\|_2 \\
    &= \sigma_{\min}(J_\theta(z)) \cdot \|\nabla_z \ell\|_2.
\end{aligned}
\end{equation}
Here, we use the property that a matrix and its transpose share the same singular values.
Finally, substituting Eq.~\ref{eq:logit-grad-norm} and the non-degeneracy assumption $\sigma_{\min}(J_\theta(z)) \ge \gamma$, we obtain:
\begin{equation}
    \|\nabla_\theta \ell\|_2 \ge \gamma \cdot (1 - p_\theta(y^*_t \mid x, y^*_{<t})).
\end{equation}
This concludes the proof.
\end{proof}

\section{Hyper parameters}
\label{app:hyper-parameters}
Regarding the hyperparameter configuration, we set the per-device batch size to 1 and employed a gradient accumulation strategy with 4 steps. 
The input sequences were truncated to a maximum length of 8,192 tokens. All models were fine-tuned for a single epoch.

During inference, we enable stochastic sampling across all models to ensure generation diversity. 
For Qwen3 and OLMo-2, we configure the hyperparameters with : do\_sample=True, temperature=0.7, top\_p=0.8, top\_k=20. 
For the Llama series, we adopt a configuration : do\_sample=True, temperature=0.6, top\_p=0.9.
Regarding generation length, we set the maximum output tokens to 32,768 for the AIME'24 benchmark to accommodate extensive reasoning, while limiting it to 8,192 for all other evaluations.

\section{Detailed Analysis of LoRA Rank across Datasets}
\label{app:rank_details}
\begin{figure*}[!t]
\centering
\includegraphics[width=\linewidth]{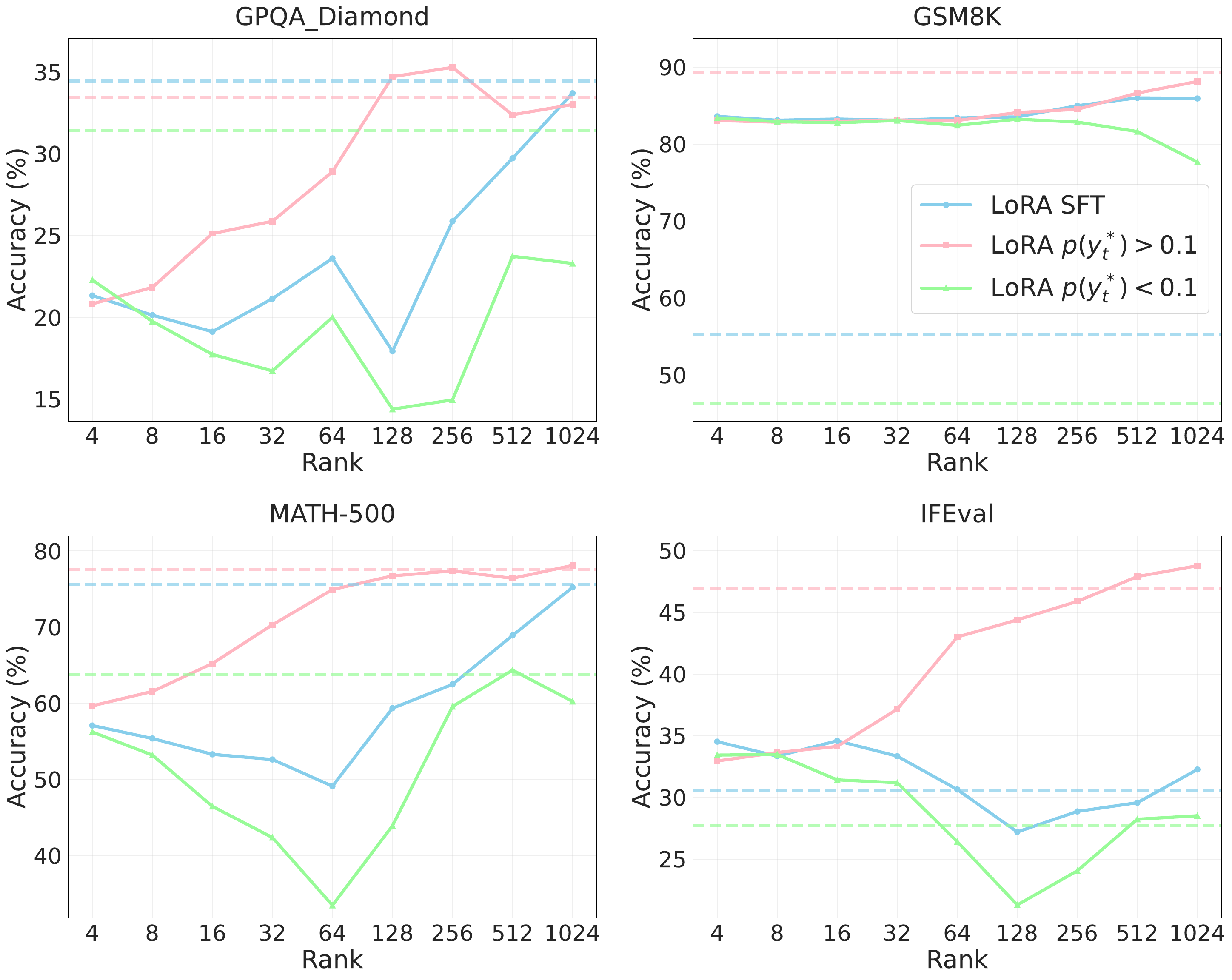}
\caption{Detailed performance comparison across different LoRA ranks for individual datasets. 
The trends corroborate our hypothesis: core tokens ($p > 0.1$) benefit from increased capacity, while non-core tokens ($p < 0.1$) induce optimization interference, particularly at high ranks.}
\label{fig:app_rank_grid}
\end{figure*}
We present the detailed performance trajectories for each dataset (GSM8K, MATH-500, GPQA-Diamond, and IFEval) across varying LoRA ranks in Figure~\ref{fig:app_rank_grid}. 
The results reveal consistent optimization patterns across diverse tasks:

\paragraph{Universal Monotonicity for Core Tokens ($p > 0.1$).}
A striking commonality across \textbf{all four datasets} is the monotonic performance growth observed when training on core tokens (pink lines). 
Whether for reasoning (GSM8K, MATH-500), knowledge recall (GPQA), or instruction following (IFEval), the model's performance steadily improves as the rank increases from 4 to 1024. 
This universality confirms a fundamental hypothesis: \textbf{core task semantics possess a high intrinsic dimensionality}. 
Consequently, providing larger parameter capacities allows the model to better resolve these critical patterns without saturation, regardless of the task type.

\paragraph{Interference from Non-Core Expressions ($p < 0.1$).}
In contrast, focusing on non-core tokens (green lines) consistently leads to suboptimal outcomes, though the manifestation varies slightly by task:
\begin{itemize}
    \item \textbf{Reasoning Tasks (GSM8K, MATH-500):} These tasks are particularly sensitive to noise. 
    At high ranks (e.g., $r=1024$), the excess capacity leads to severe overfitting on non-core expressions, causing a sharp performance drop.
    \item \textbf{Knowledge \& Instruction Tasks (GPQA, IFEval):} The standard SFT baseline (blue lines) and non-core token training often exhibit fluctuations or stagnation at medium-to-high ranks, struggling to maintain the upward trajectory seen in the core token setting.
    This further highlights that filtering out non-core expressions is essential for stable scaling.
\end{itemize}

\section{Detailed Analysis of Training Dynamics across Epochs}
\label{app:epoch_details}

We further examine the training stability and convergence speed on individual datasets, as illustrated in Figure~\ref{fig:app_epoch_grid}. 
The trajectories provide granular evidence for the efficiency of our method:

\begin{figure*}[!t]
\centering
\includegraphics[width=\linewidth]{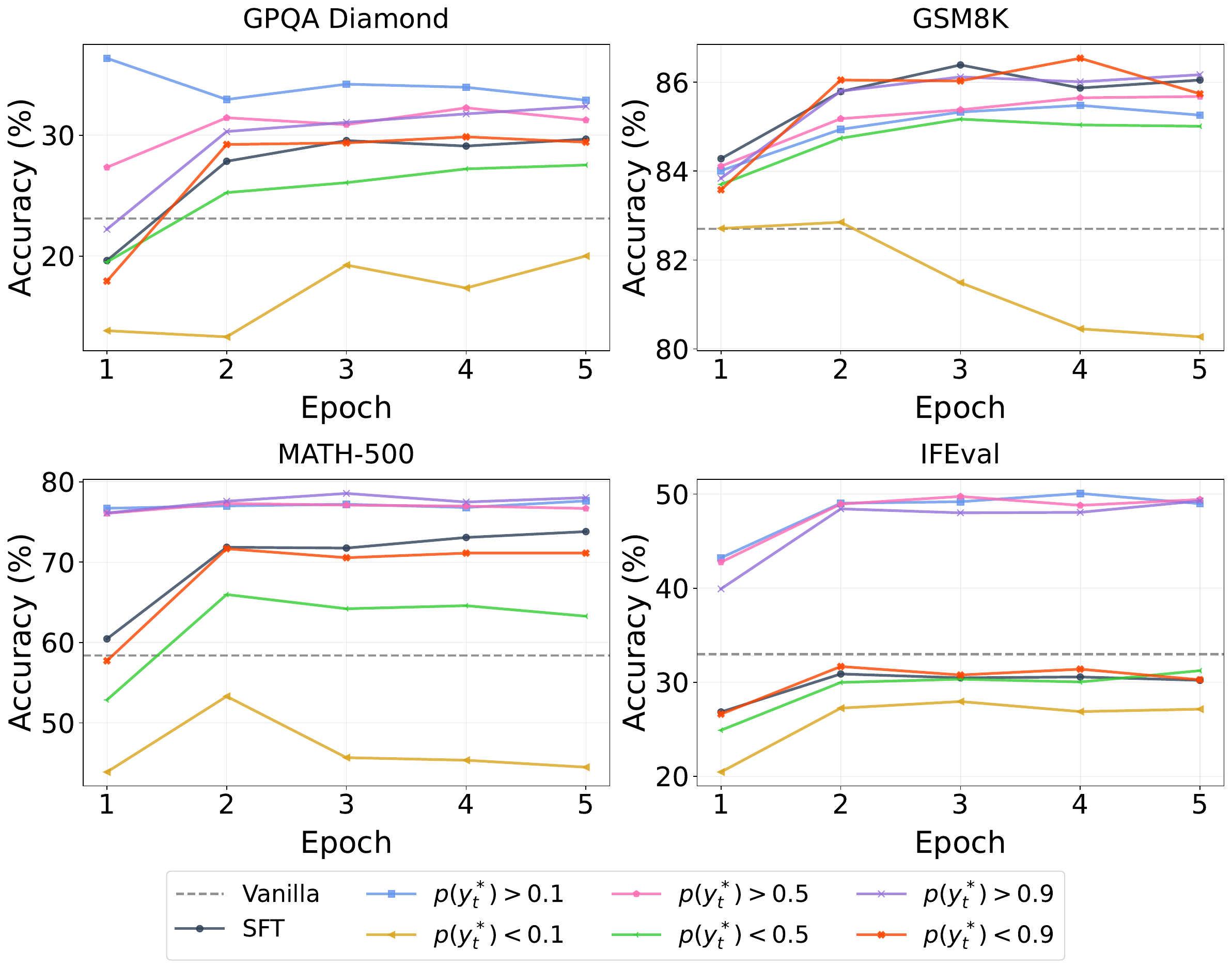}
\caption{Performance evolution across training epochs for individual datasets. 
ProFit ($p > \tau$) exhibits rapid convergence and stability, whereas training on low-probability tokens ($p < \tau$) suffers from instability and overfitting to non-core expressions.}
\label{fig:app_epoch_grid}
\end{figure*}

\paragraph{Rapid Convergence and Stability.}
Across all benchmarks, ProFit settings (solid lines, $p > \tau$) demonstrate superior convergence efficiency, typically reaching near-optimal performance within just 2 epochs and maintaining stability throughout the training process. 
In contrast, the SFT baseline often requires more steps to plateau or exhibits fluctuations.

\paragraph{Overfitting to Non-Core Expressions.}
The risks of training on low-probability tokens are clearly visible in reasoning tasks. 
For instance, in GSM8K and MATH-500, the performance of the $p < 0.1$ setting (yellow line) peaks early but subsequently degrades as training progresses (Epoch 3-5). 
This \textit{inverted-U} pattern strongly suggests that prolonged exposure to low-probability tokens leads the model to overfit to non-core expressions , thereby impairing its underlying reasoning logic.

\paragraph{The Gap in Instruction Following.}
In IFEval, a distinct performance chasm is observed: models trained on high-probability tokens stabilize around 50\% accuracy, whereas those focused on low-probability tokens stagnate below 30\%. 
This indicates that the core semantics required for instruction following are almost exclusively encoded in high-probability tokens, while low-probability tokens contribute little to this capability.

\section{Case Study}
As shown in Table~\ref{tab:case_study_poly}, SFT succumbs to logical hallucinations by ignoring key interaction terms, whereas ProFit maintains a coherent reasoning chain to derive the correct solution. 
This demonstrates that masking non-core expressions effectively safeguards the model's core logic against superficial errors.

\begin{figure*}[!t]
    \centering
    \includegraphics[width=\textwidth]{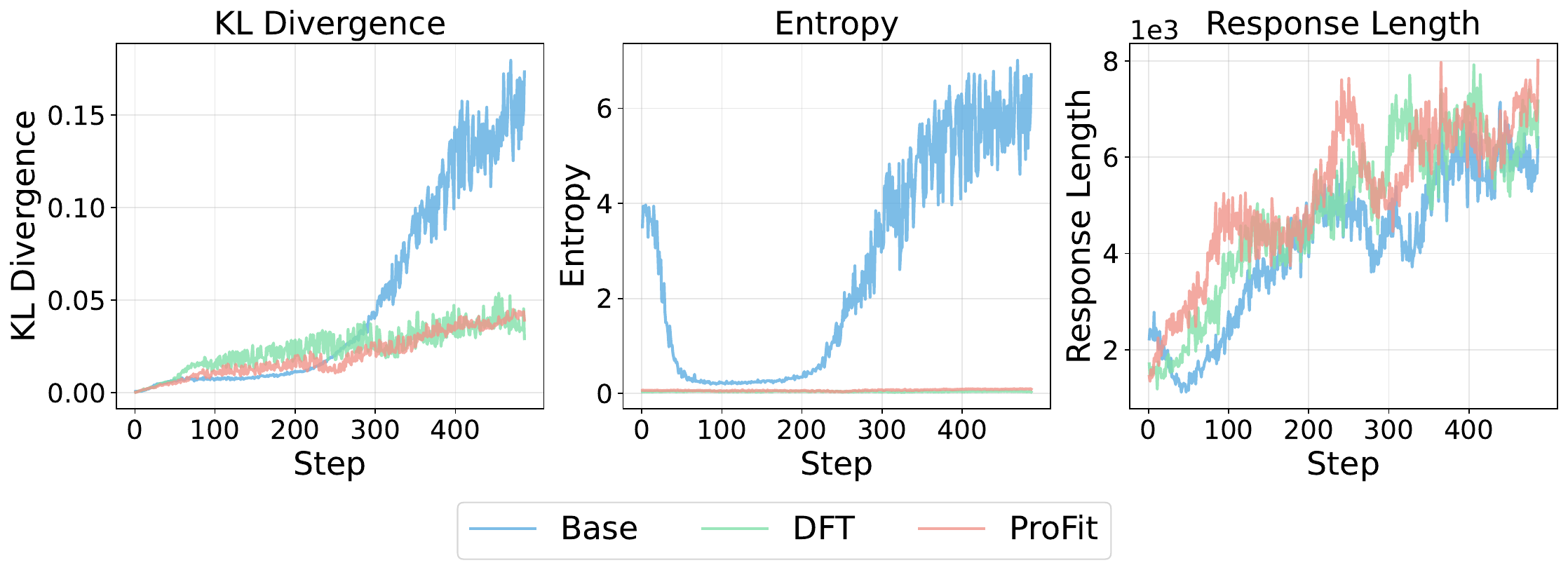}    
    \caption{Training Dynamics in RL Stage. We compare the KL Divergence, Entropy, and Response Length of models initialized with Base, DFT, and ProFit strategies. ProFit demonstrates superior stability (low KL), confident convergence (low Entropy), and evolves deeper reasoning capabilities (highest Response Length).}
    \label{fig:rl_dynamics}
\end{figure*}

\section{Benchmark Details}
\label{app:benchmark_details}

We evaluate our method on diverse benchmarks covering reasoning and instruction following capabilities:

\begin{itemize}
    \item \textbf{GPQA-Diamond}~\citep{rein2024gpqa}: A dataset of 198 expert-level, \textit{Google-proof} science questions testing advanced reasoning.
    \item \textbf{IFEval}~\citep{zhou2023instruction}: Evaluates the model's ability to follow verifiable formatting instructions and constraints.
    \item \textbf{GSM8K}~\citep{cobbe2021training}: A classic benchmark consisting of grade-school math word problems requiring multi-step reasoning.
    \item \textbf{MATH-500}~\citep{lightman2023let}: A challenging subset of 500 competition-level mathematics problems from the Minerva dataset.
    \item \textbf{AIME'24}~\citep{aime2024dataset}: A set of high-difficulty problems from the 2024 American Invitational Mathematics Examination, testing frontier mathematical capabilities.
    \item \textbf{Minerva}~\citep{hendrycks2021measuring}: A comprehensive dataset of 12,500 challenging competition mathematics problems ranging from pre-algebra to calculus, serving as a standard benchmark for mathematical reasoning.
    \item \textbf{OlympiadBench}~\citep{he2024olympiadbench}: A large-scale, bilingual, and multimodal benchmark featuring Olympiad-level problems in mathematics and physics, designed to evaluate AGI capabilities in complex scientific reasoning.
    \item \textbf{HellaSwag}~\citep{zellers2019hellaswag}: A benchmark for evaluating commonsense natural language inference, requiring models to choose the most logical and natural continuation of a given text snippet.
    \item \textbf{ARC-c}~\citep{clark2018think}: The Challenge set of the AI2 Reasoning Challenge, consisting of difficult grade-school science questions that require advanced logic and reasoning beyond simple information retrieval.
    \item \textbf{HumanEval}~\citep{chen2021evaluating}: A benchmark designed to evaluate code generation capabilities, featuring programming problems (primarily in Python) that test algorithmic logic and basic coding skills.
    \item \textbf{MMLU-Pro}~\citep{wang2024mmlu}: An enhanced, more challenging version of the Massive Multitask Language Understanding (MMLU) benchmark. In our evaluation, we specifically report performance on the \textbf{History} and \textbf{Chemistry} subsets to assess advanced, domain-specific knowledge and reasoning.
\end{itemize}

\section{Training Dynamics Analysis in RL Stage}
\label{app:appendix_rl_dynamics}

To further investigate the impact of different initialization strategies on the reinforcement learning (RL) process, we visualized the training dynamics of three key metrics: KL Divergence, Entropy, and Response Length. The comparative results are presented in Figure \ref{fig:rl_dynamics}.

\paragraph{KL Divergence Stability.} 
As shown in the left panel of Figure \ref{fig:rl_dynamics}, the Base model (blue line) exhibits a rapid and uncontrolled increase in KL divergence, reaching approximately 0.17 by the end of training. 
This sharp rise suggests that without a robust SFT warm-up, the policy drifts significantly from the reference model, potentially leading to reward hacking or language degeneration. 
In contrast, \textbf{ProFit} (red line) and DFT (green line) maintain a remarkably low and stable KL divergence (staying below 0.05). 
This indicates that ProFit effectively constrains the policy update within a safe trust region, ensuring that the model improves its mathematical reasoning capabilities while preserving its general linguistic coherence.

\paragraph{Entropy.} 
The entropy curves reveal distinct convergence behaviors. 
The Base model demonstrates chaotic behavior, where entropy initially collapses and then drastically rebounds to extremely high values ($>6.0$), indicating a failure to converge to a stable policy and likely degenerating into generating high-randomness noise.
Conversely, \textbf{ProFit} maintains a consistently low entropy (ending around 0.09), similar to DFT. 
This low entropy signifies high confidence in the generated reasoning paths. 
Notably, unlike DFT which saturates near 0.03, ProFit maintains a slightly higher entropy margin, suggesting it retains a healthy level of exploration potential while remaining focused on high-reward solutions.

\paragraph{Response Length.} 
The most significant differentiator lies in the response length, which serves as a proxy for the depth of Chain-of-Thought (CoT) reasoning. 
While all models show an increasing trend in response length, \textbf{ProFit} demonstrates the most robust growth pattern. 
Starting from $\sim$1,400 tokens, ProFit rapidly learns to expand its reasoning steps, surpassing the Base model around step 150 and eventually achieving the highest average response length of over \textbf{8,000 tokens} at the end. 
Compared to DFT (ending at 7,100 tokens) and Base (6,400 tokens), ProFit's superior length indicates that the probability-guided initialization encourages the RL algorithm to explore deeper and more complex reasoning chains. 
This aligns perfectly with our main results, where ProFit excels in complex tasks like OlympiadBench, which require extensive multi-step deductions.

\begin{table*}[p]  
\small
\begin{tabular}{lp{0.85\textwidth}}
\toprule
\textbf{Question} & 
The polynomial $x^3 - 3x^2 + 4x - 1$ is a factor of $x^9 + px^6 + qx^3 + r.$ Enter the ordered triple $(p,q,r).$
Please reason step by step, and put your final answer within \boxed{}.
\\
\midrule 
\textbf{Reference} & 
Let $\alpha$ be a root of $x^3 - 3x^2 + 4x - 1 = 0,$ so $\alpha^3 = 3 \alpha^2 - 4 \alpha + 1.$
Substituting this relation repeatedly into higher powers:
$\alpha^9 = -63 \alpha^2 + 190 \alpha - 54.$
Then $\alpha^9 + p \alpha^6 + q \alpha^3 + r = (-5p + 3q - 63) \alpha^2 + (-11p - 4q + 190) \alpha + (4p + q + r - 54).$
For this to be 0 for all roots, we set coefficients to 0:
$-5p + 3q = 63, \quad 11p + 4q = 190, \quad 4p + q + r = 54.$
Solving this system yields $(p,q,r) = \boxed{(6,31,-1)}.$
\\
\midrule
\textbf{SFT} & 
To determine the ordered triple $(p, q, r)$, we assume $x^9 + px^6 + qx^3 + r = (x^3 - 3x^2 + 4x - 1) \cdot g(x)$, where $g(x) = ax^6 + bx^5 + cx^4 + dx^3 + ex^2 + fx + g$.
\newline
\textbf{Step 4: Match coefficients}
\newline
Let's expand the product step-by-step:
\begin{itemize}
    \item The $x^9$ term: $x^3 \cdot ax^6 = ax^9 \implies \boldsymbol{a = 1}$.
    \item The $x^8$ term: $x^3 \cdot bx^5 = bx^8 \implies \boldsymbol{b = 0}$ \textbf{\color{red}{(Incorrectly ignores interaction terms)}}.
    \item The $x^7$ term: $x^3 \cdot cx^4 = cx^7 \implies \boldsymbol{c = 0}$.
    \item The $x^6$ term: Combining terms yields $(d - 3)x^6 - cx^4$. We match this to $p$: $\boldsymbol{d - 3 = p}$ and $c = 0$.
    \item The $x^5$ term: Combining terms yields $ex^5 - dx^3$. We need this to match 0: $\boldsymbol{e = 0}$ and $\boldsymbol{d = 0}$.
    \item The $x^4$ term: Combining terms yields $fx^4$. We need this to match 0: $\boldsymbol{f = 0}$.
    \item The constant term: $-g = r \implies \boldsymbol{g = r}$.
\end{itemize}
\textbf{Step 5: Combine the results}
\newline
From the above steps, we have: $a = 1, b = 0, c = 0, d = 0, e = 0, f = 0, g = 0, r = 0$.
Thus:
$p = d - 3 = 0 - 3 = -3$,
$q = g = 0$,
$r = 0$.
\newline
\textbf{Final Answer:}
$\boxed{(-3, 0, 0)}$
\\
\midrule
\textbf{ProFit} & 
To determine the ordered triple $(p, q, r)$, we denote $f(x) = x^3 - 3x^2 + 4x - 1$. We assume $g(x) = x^6 + ax^5 + bx^4 + cx^3 + dx^2 + ex + f$.
Then: $(x^3 - 3x^2 + 4x - 1)(x^6 + ax^5 + ... + f) = x^9 + px^6 + qx^3 + r$.
\newline
\textbf{Step-by-step Coefficient Comparison:}
\newline
Expanding the LHS and comparing with $x^9 + px^6 + qx^3 + r$:
\begin{enumerate}
    \item $x^8$ coeff: $a-3 = 0 \implies \boldsymbol{a = 3}$.
    \item $x^7$ coeff: $b-3a+4 = 0 \implies b-9+4=0 \implies \boldsymbol{b = 5}$.
    \item $x^6$ coeff: $c-3b+4a-1 = p \implies c-15+12-1 = p \implies \boldsymbol{p = c-4}$.
    \item $x^5$ coeff: $d-3c+4b-1a = 0 \implies d-3c+20-3 = 0 \implies \boldsymbol{d = 3c-17}$.
    \item $x^4$ coeff: $e-3d+4c-1b = 0 \implies e-3(3c-17)+4c-5 = 0 \implies \boldsymbol{e = 5c-46}$.
    \item $x^3$ coeff: $f-3e+4d-1c = q \implies f-3(5c-46)+4(3c-17)-c = q \implies \boldsymbol{q = f-4c+70}$.
    \item $x^2$ coeff: $-3f+4e-1d = 0 \implies ... \implies f = \frac{17c-167}{3}$.
    \item $x^1$ coeff: $-f+4e-1d = 0 \implies ... \implies 34c-340 = 0 \implies \boldsymbol{c = 10}$.
    \item Constant: $-f = r \implies r = -1$.
\end{enumerate}
\textbf{Final Calculation:}
Now, substituting $c = 10$ back into the expressions:
$p = c - 4 = 10 - 4 = 6$.
$q = f - 4c + 70 = 1 - 40 + 70 = 31$.
$r = -1$.
\newline
Thus, the ordered triple is: $\boxed{(6, 31, -1)}$
\\
\bottomrule
\end{tabular}
\caption{\textbf{Comparison of model outputs on the polynomial factorization task.} SFT hallucinates constraints (e.g., setting $b=0$ and $c=0$ independently), leading to an incorrect solution. ProFit maintains a coherent logical chain throughout the coefficient comparison, deriving the correct ordered triple.}
\label{tab:case_study_poly}
\end{table*}

\begin{figure*}[p]
\centering
\includegraphics[width=\linewidth]{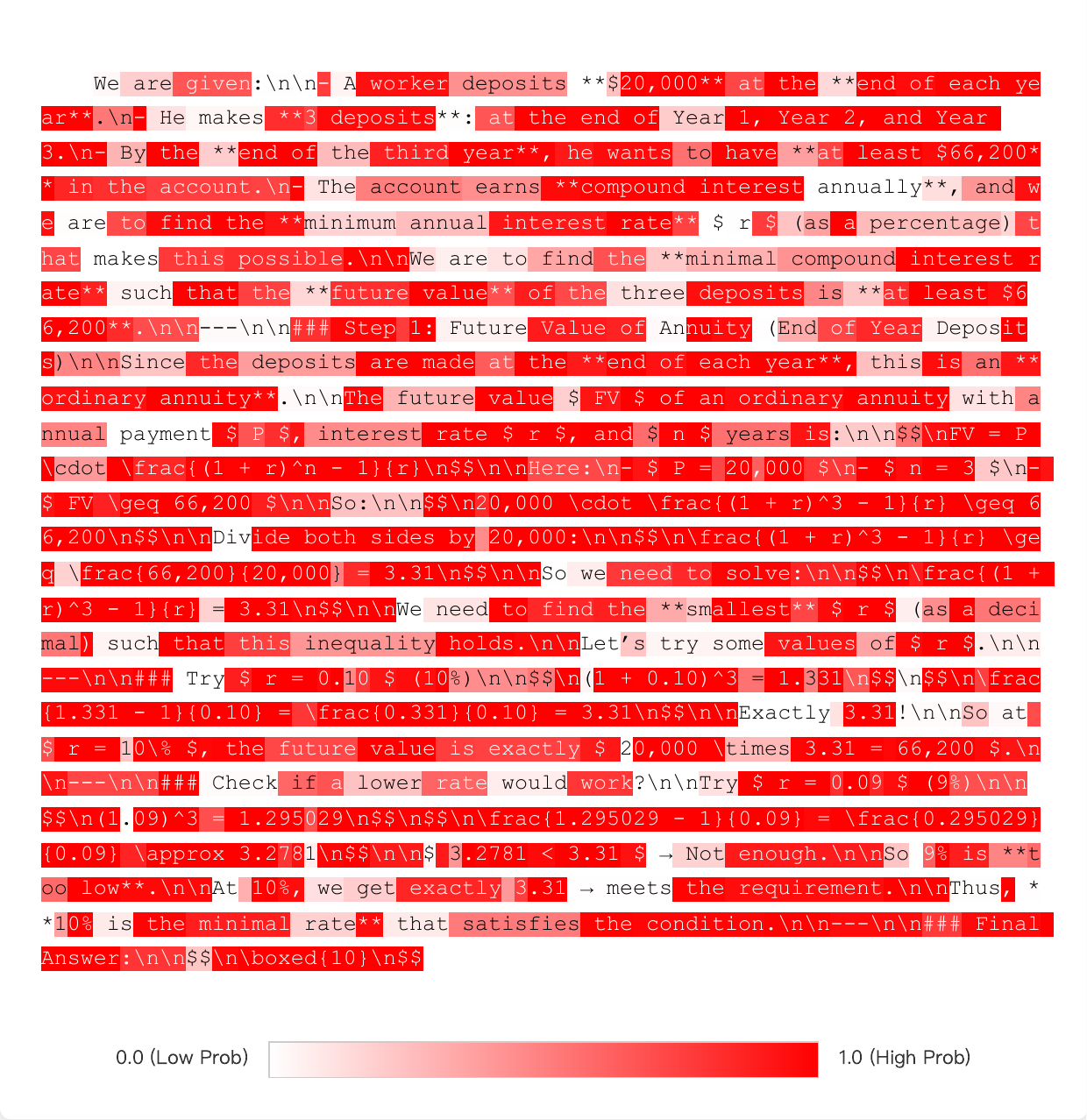}
\caption{Visualization of token probability}
\label{fig:token_prob}
\end{figure*}

\end{document}